\documentclass[sigconf,natbib=true]{acmart}

\usepackage[inline, shortlabels]{enumitem}
\usepackage{graphicx}
\usepackage{wrapfig}
\usepackage{multirow}
\usepackage{cleveref}
\AtBeginDocument{%
  }

\setcopyright{acmlicensed}
\copyrightyear{2018}
\acmYear{2018}
\acmDOI{XXXXXXX.XXXXXXX}
\acmConference[Conference acronym 'XX]{Make sure to enter the correct
  conference title from your rights confirmation email}{June 03--05,
  2018}{Woodstock, NY}
\acmISBN{978-1-4503-XXXX-X/2018/06}

\crefformat{section}{\S#2#1#3}
\crefformat{subsection}{\S#2#1#3}
\crefformat{subsubsection}{\S#2#1#3}
\crefrangeformat{section}{\S\S#3#1#4 to~#5#2#6}
\crefmultiformat{section}{\S\S#2#1#3}{ and~#2#1#3}{, #2#1#3}{ and~#2#1#3}

\copyrightyear{2025}
\acmYear{2025}
\setcopyright{acmlicensed}\acmConference[SIGIR '25]{Proceedings of the 48th
International ACM SIGIR Conference on Research and Development in
Information Retrieval}{July 13--18, 2025}{Padua, Italy}
\acmBooktitle{Proceedings of the 48th International ACM SIGIR Conference on
Research and Development in Information Retrieval (SIGIR '25), July 13--18,
2025, Padua, Italy}
\acmDOI{10.1145/3726302.3729901}
\acmISBN{979-8-4007-1592-1/2025/07}

\begin{document}


\title{BALI: Enhancing Biomedical Language Representations through Knowledge Graph and Language Model Alignment}

\author{Andrey Sakhovskiy}
\orcid{0000-0003-2762-2910}
\affiliation{%
  \institution{Sber AI}
  \institution{Skoltech}
  \city{Moscow}
  \country{Russia}
}
\email{andrey.sakhovskiy@gmail.com}

\author{Elena Tutubalina}
\orcid{0000-0001-7936-0284}
\affiliation{%
  \institution{AIRI}
  \institution{Sber AI}
  \institution{ISP RAS Research Center for Trusted AI}
  \city{Moscow}
  \country{Russia}
}
\email{tutubalinaev@gmail.com}

\begin{abstract}
  In recent years, there has been substantial progress in using pretrained Language Models (LMs) on a range of tasks aimed at improving the understanding of biomedical texts. Nonetheless, existing biomedical LLMs show limited comprehension of complex, domain-specific concept structures and the factual information encoded in biomedical Knowledge Graphs (KGs). In this work, we propose \textbf{BALI} (\textbf{B}iomedical Knowledge Graph and Language Model \textbf{Ali}gnment), a novel joint LM and KG pre-training method that augments an LM with external knowledge by the simultaneous learning of a dedicated KG encoder and aligning the representations of both the LM and the graph. For a given textual sequence, we link biomedical concept mentions to the Unified Medical Language System (UMLS) KG and utilize local KG subgraphs as cross-modal positive samples for these mentions. Our empirical findings indicate that implementing our method on several leading biomedical LMs, such as PubMedBERT and BioLinkBERT, improves their performance on a range of language understanding tasks and the quality of entity representations, even with minimal pre-training on a small alignment dataset sourced from PubMed scientific abstracts.

\end{abstract}




\maketitle

\section{Introduction}

In recent years, advancements in biomedical Natural Language Processing (NLP) have been largely driven by the development of domain-specific pre-trained Language Models (LMs)~\citep{alsentzer-etal-clinicalBERT-2019,beltagy-etal-2019-scibert,MichalopoulosUmlsBERTWK21,DBLP:conf/acl/YasunagaLL22,DBLP:journals/health/GuTCLULNGP22,DBLP:conf/acl-clinicalnlp/MannionSG23,sakhovskiy-etal-2024-biomedical}. Despite the recent success of Large Language Models (LLMs) in the general domain, they fall short of lightweight domain-specific biomedical LMs~\citep{DBLP:journals/health/GuTCLULNGP22,DBLP:conf/acl/YasunagaLL22} by a large margin~\citep{ChatGPTOnBLURB,bai2024kgquiz}. While domain-specific models have shown remarkable performance on biomedical NLP benchmarks, for instance, on the Biomedical Language Understanding and Reasoning Benchmark (BLURB)~\cite{DBLP:journals/health/GuTCLULNGP22}, they have been shown to impose limited domain-specific factual knowledge understanding~\citep{BioLamaDataset2021,MedLamaDataset2022} and information extraction~\citep{ChatGPTOnBLURB}. 

The concept structure and factual knowledge within a specific domain are often represented through extensive knowledge graphs (KGs), which can describe millions of domain-specific concepts and their inter-relations. A notable example in the biomedical domain is the Unified Medical Language System (UMLS)\footnote{\url{https://www.nlm.nih.gov/research/umls/index.html}} KG~\citep{Bodenreider_UMLS2004}, a comprehensive meta-thesaurus covering over 4M concept from 166 lexicons/thesauri. Recent lines of research have iteratively improved the current state-of-the-art performance on biomedical entity representations by pre-training either on UMLS concept names~\citep{SapBERTMonoLingual2022,SapBERTMultilingual,DBLP:journals/jbi/YuanZSLWY22} or aligned text-KG subgraph pairs~\citep{sakhovskiy-etal-2024-biomedical}. However, these work mostly fine-tuned LMs for entity linking, limiting their applicability beyond this specific task. This narrow focus can hinder the models' ability to generalize across diverse biomedical texts and concepts.

Recent efforts to improve the knowledge capabilities of LMs involve integrating text and knowledge graphs (KGs) in a shallow or one-way manner \citep{zhang2019ernie,wang2021kepler,sun2021ernie,baek2023knowledge} (e.g., from KG to text for retrieval-augmented methods like RAG \citep{lewis2020retrieval}, REALM \cite{guu2020retrieval}, and REPLUG \citep{shi2024replug}), which could hinder multi-hop reasoning. Another approach is using an interaction token  \citep{DBLP:conf/iclr/0001BYRLML22,DBLP:conf/nips/YasunagaBR0MLL22} or a projector \cite{tian2024graph} that depends on implicit exchanges between modalities. Unlike previous efforts, we explore the alignment of the uni-modal embedding spaces using anchors to better capture interconnected information and dependencies between textual and graph modalities.  This alignment may contribute to enhanced multi-hop reasoning capabilities, as the model can more effectively traverse and reason across the aligned spaces.

\begin{figure*}[t]
  \centering
  \includegraphics[width=\linewidth]{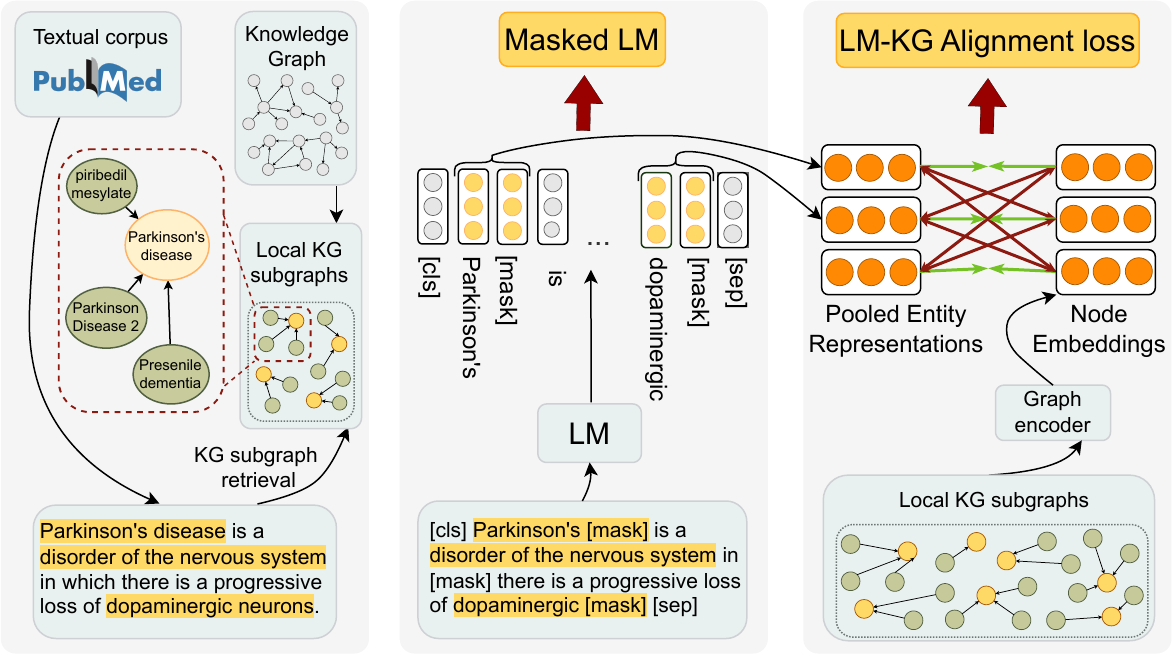}
  \caption{The overall framework. We first retrieve subgraphs from the knowledge graph based on the entities in a text fragment (\cref{sec:notation}). We then develop \textbf{BALI} (Knowledge \textbf{Gra}ph and \textbf{B}iomedical Language Model A\textbf{li}gnment) to align knowledge between a textual encoder and a graph encoder (\cref{sec:repres}).  We utilize two objectives: (1) masked language modeling (MLM), which masks some tokens in the input text and then predicts them, and (2) cross-modal alignment, which pull two representations of a concept closer in a combined embedding space. Since the entity representation is pooled over a textual sequence masked for MLM objective, this alignment objective further enforces LM to infer relevant information from the whole sequence (\cref{sec:obj}).
}\label{fig:teaser} %
\end{figure*}

In this paper, we introduce Knowledge \textbf{Gra}ph and \textbf{B}iomedical Language Model A\textbf{li}gnment (BALI), a novel pre-training approach that enhances LM with external knowledge by concurrently training a distinct KG encoder and aligning the representations of both the LM and the graph. Specifically, as in Figure \ref{fig:teaser}, given a (text, local KG) pair, a graph neural network (GNN) is utilized to capture and encode the graph knowledge into node embeddings, while pre-trained LM is used to obtain textual entity representations. Textual entity representations and concept node representations are used as anchors to align the two uni-modal embedding spaces. 
In this work, we seek to answer the following research questions (RQs): 
\begin{enumerate}[start=1,label={\bfseries RQ\arabic*:}]
\item Is the proposed cross-modal LM-KG alignment procedure with explicit alignment between two representation spaces beneficial for biomedical NLP downstream tasks?
\item What is the most effective graph representation for LM-KG alignment?
\item Is the utilization of an external graph encoder more effective for cross-modal LM-KG alignment or using graph linearization followed by LM encoding is sufficient?
\end{enumerate}

To comprehensively assess our model, we perform extensive experiments across several benchmarks for question answering and entity linking tasks.  
Initially, we pretrain several LMs with BALI, leveraging the PubMed corpus and UMLS KG. Our experiments demonstrate that BALI outperforms several biomedical language models, including BioLinkBERT~\citep{DBLP:conf/acl/YasunagaLL22} and PubMedBERT~\citep{DBLP:journals/health/GuTCLULNGP22}. Specifically, PubMedBERT shows mean accuracy improvements of 2.1\%, 1.7\%, and 6.2\% on the PubMedQA, MedQA, and BioASQ benchmarks, respectively. BALI significantly enhances the ability of LMs to generate distinguishable and informative representations of biomedical concepts. In particular, BioLinkBERT with BALI pretraining performs on par or slightly better than the task-specific SapBERT model, which is pre-trained on 12M UMLS triples (4M concept nodes). Our research highlights that our cross-modal knowledge graph alignment, applied to both text and the knowledge graph, notably enhances language-knowledge representations after a small pre-trainning stage involving 1.5M sentences and 600K nodes only. The source code as well as pre-trained models are publicly available at: \url{https://github.com/Andoree/BALI-BERT}.










\section{Related Work}










\paragraph{Knowledge-Augmented Language Models}

One line of research on knowledge-enhanced LMs~\citep{DBLP:conf/aaai/LiuZ0WJD020,DBLP:conf/coling/SunSQGHHZ20,DBLP:conf/acl/KeJRCWSZH21,DBLP:conf/acl-clinicalnlp/MannionSG23, DBLP:journals/jbi/YuanZSLWY22,DBLP:conf/naacl/MoiseevDAJ22} attempted to infuse factual information  into LM input either by augmenting natural language texts with relational triples or directly training on relational triples. Various methods~\citep{sevgili-etal-2019-improving,DBLP:conf/acl/ZhangHLJSL19, DBLP:conf/emnlp/HeZXJLYX20,DBLP:journals/tacl/WangGZZLLT21,DBLP:conf/emnlp/PetersNLSJSS19,DBLP:conf/chiir/SchildwachterBZ19,DBLP:journals/corr/abs-2007-00655,DBLP:conf/aaai/Yu0Y022,DBLP:conf/naacl/KangBH22} augment in-context entity representation with external knowledge retrieved from KG. While such methods are able to improve quality on NLP tasks, they usually perform unidirectional information fusion for improved LM embeddings using either a single LM for both modalities or static KG node embeddings. Static node embeddings are unable to capture node semantics and only capture structural information, Transformer-based~\citep{DBLP:conf/nips/VaswaniSPUJGKP17} LM's architecture is inherently dense which confronts the sparse nature of KGs. Recently proposed GreaseLM~\citep{DBLP:conf/iclr/0001BYRLML22} and DRAGON~\citep{DBLP:conf/nips/YasunagaBR0MLL22} models improve LM reasoning ability by introducing bidirectional cross-modal interaction text and grounded KG subgraph interaction through specialized cross-modal LM token for enhanced question answering.
However, both models depend on implicit intermodal exchanges: the LM accesses KG information via a single token initialized with pooled subgraph representation, while the graph encoder receives semantic input through an interaction node initialized with pooled sentence representation. 
Meanwhile, these modalities offer complementary representations of a single entity, capturing different contexts: sentences for the LM and KG subgraphs for the graph encoder implying that the two uni-modal spaces can be aligned through entities serving as anchors in a unified embedding space.

Recently proposed method~\cite{tian2024graph} encodes subgraphs based on the entities present in the question and options.
In contrast with direct feeding of KG triples into LLMs \cite{baek2023knowledge}, this approach utilizes a GNN, a cross-modality pooling module, and a domain projector to send the encoded subgraphs to LLMs for inference, alongside the input text embeddings. This represents an alternative prompt-based direction focusing on parameter-efficient fine-tuning.












\paragraph{Graph Representation Learning}
A series of translation-based node representation methods~\citep{DBLP:journals/corr/YangYHGD14a,DBLP:conf/nips/BordesUGWY13,DBLP:conf/icml/TrouillonWRGB16,kutuzov-etal-2019-learning,DBLP:conf/nips/Kazemi018,DBLP:conf/iclr/SunDNT19} models a relation triplet (graph edge) as a translation between head and tail nodes. Initially, these methods learned static node embedding matrix as well as relation embeddings via the link prediction task contrastively with knowledge triples present in a KG being positive samples and non-present ones being negative samples. Experimental evaluation of translation-based methods for biomedical concept representation~\citep{DBLP:conf/bionlp/ChangBACBT20} indicates that these methods fall short of the LM-based approach due to a lack of essential semantical information present in texts. While translation-based methods model each edge individually, Message Passing (MP)~\citep{DBLP:conf/icml/GilmerSRVD17} graph neural networks obtain node embeddings by passing and aggregating messages from multiple neighboring nodes at once. Various architectures under the MP framework mostly differ in message aggregation function. For instance, GraphSAGE~\citep{DBLP:conf/nips/HamiltonYL17} performs mean-pooling over neighboring nodes, and Graph Attention Network (GAT)~\citep{DBLP:conf/iclr/VelickovicCCRLB18,DBLP:conf/iclr/Brody0Y22}  applies an attention-based aggregation. In our work, we adopt GAT for local KG subgraph aggregation as it has proved itself an effective  graph encoder for LM-KG interaction applications~\citep{DBLP:conf/naacl/YasunagaRBLL21,DBLP:conf/iclr/0001BYRLML22,DBLP:conf/nips/YasunagaBR0MLL22,DBLP:conf/clef/SakhovskiySKT23,sakhovskiy-etal-2024-biomedical}. Another approach~\citep{wang2021kepler,DBLP:conf/paclic/SalnikovLRNBMP23} gets rid of additional memory footprint introduced by an external graph encoder by linearizing KG subgraphs into textual strings encoded with an LM.



\paragraph{Cross-Modal Alignment} Our research is inspired by recent advancements in aligning multiple uni-modal representations across various domains. \citet{DBLP:conf/nips/KohFS23,DBLP:conf/icml/KohSF23} trains a small alignment network to align images with their captions for cross-modal visual and textual generative tasks.~\citet{DBLP:conf/nips/LiuLWL23a} learns a lightweight projection to align visual and textual features for improved multimodal image and language understanding.
\citet{DBLP:conf/acl/KeJRCWSZH21} introduced a method to align entities in text with their representations in graphs, improving graph summarization. Unlike prior work, we perform explicit cross-modal alignment by directly minimizing distances between cross-modal paired representations of a single biomedical concept.

\paragraph{Large Language Models} Despite the recent remarkable success of large language models (LLMs) in a wide range of NLP tasks~\citep{GPT4-technical-report,ChatGPTOnBLURB,DBLP:journals/corr/abs-2302-13971,dale-etal-2021-text,li2024developingchatgptbiology}, they exhibit a few limitations that hinders them from becoming a universal tool for biomedical NLP tasks~\citep{ChatGPTOnBLURB}. First, LLMs are resource-intensive and have billions of parameters compared to lightweight biomedical BERT-based encoders~\citep{DBLP:conf/naacl/DevlinCLT19} with only a few million parameters~\citep{DBLP:conf/acl/YasunagaLL22,DBLP:journals/health/GuTCLULNGP22}. Second, even though LLMs exhibit superior performance in a wide range of tasks, especially in language generation tasks, recent research has proved domain-specific BERT models to have higher quality in biomedical information retrieval tasks~\citep{ChatGPTOnBLURB}. 





\section{Problem Statement/Notation} \label{sec:notation} 

\paragraph{Biomedical Knowledge Graph} 




Formally, a Knowledge Graph can be defined as $\mathcal{G} = (\mathcal{V}, \mathcal{E}, \mathcal{R})$, where $\mathcal{V}$ is the set of biomedical concepts, $\mathcal{E} \subset \mathcal{V} \times \mathcal{R} \times \mathcal{V}$ is the set of labeled edges, and $\mathcal{R}$ are possible relation types. In UMLS, one of the largest biomedical KGs, a node $v \in \mathcal{V}$ can be represented with a set of $k \geq 1$ distinct synonymous concept names $S_v = \{s^v_1, s^v_2,\dots, s^v_k\}$. Thus, a concept $v \in \mathcal{V}$ can be represented with two complementary modalities: (i) a textual modality described by $S_v$, (ii) a KG modality expressed with local subgraph $\mathcal{G}_v \subset \mathcal{G}$ centered around $v$. Additionally, textual concept representations can be learnt from raw texts they are mentioned in.  


\paragraph{KG Subgraphs} 
From KG perspective, a node $v \in \mathcal{V}$ can be described by the structure of its local KG subgraph, denoted as $\mathcal{G}_v = (\mathcal{V}_v, \mathcal{E}_v, \mathcal{R}) \subset \mathcal{G}$, consisting of 1-hop neighbors subgraph centered around  $v$: $\mathcal{E}_v = \{(u, r, v) \in \mathcal{E}\}, \mathcal{V}_v = \{u\,\,|\,\,(u, r, v) \in \mathcal{E}_v \} \cup v$. Here, $\mathcal{G}_v$ can be viewed as a structural KG-induced context for a concept. 


\paragraph{Alignment Intuition}

While graph encoder $GNN$ can capture the hierarchy of in-domain concepts along with other inter-concept relationships, textual encoder $LM$ can provide deeper insights into concept semantics learnt from raw texts. Conversely, $LM$ may struggle to effectively learn the intricate concept structure from texts alone. Thus, we assume that two embedded representations $\bar{g}_v$ and $\bar{e}_v$ are complementary representations encoding different features of the same concept $v$. Our goal is to align these two uni-modal entity representations by enabling a mutual knowledge exchange. Since we assume $\bar{e}_v$ and $\bar{g}_v$ to be complementary representations capturing different features of a concept $v$, we propose to use these embeddings as anchors for aligning inner representations of $GNN$ and $LM$.

\section{Methodology}

Overall, our objective is to align the knowledge between a textual encoder and a graph encoder using textual entity representations and concept node representations as anchors for aligning two uni-modal embedding spaces. Compared to the existing joint LM-KG methods~\citep{DBLP:conf/naacl/YasunagaRBLL21,DBLP:conf/nips/YasunagaBR0MLL22}, our BALI method strives to limit the interaction of modalities to the pre-training stage only completely eliminating the need for inference-time entity linking and KG subgraphs retrieval.

\subsection{Uni-Modal Representations} \label{sec:repres}



\paragraph{Entity Representations}
Let $T = (t_1, t_2,\dots,t_N)$ denote a textual sequence  consisting of $N$ tokens. 
To encode the sequence, we adopt a language model $LM$ that is based on Transformer encoder~\citep{DBLP:conf/nips/VaswaniSPUJGKP17}: 
$$
H_T = (\bar{h}_1, \bar{h}_2,\dots,\bar{h}_N) = LM\{(t_1, t_2,\dots,t_N)\},
$$
where $\bar{h}_j \in \mathbb{R}^d$ is a $d$-dimensional embedding for $j$-th token in the sequence. Here, $H_T \in \mathbb{R}^{N \times d}$ is a textual embedding matrix for a sequence $T$. We assume that the text $T$ mentions $M$ KG nodes, denoted as $V_T = \{v_i\}^M_{i=1} \subset \mathcal{V}$. For each concept $v \in \mathcal{V}_T$ there is a subset of tokens from $T$ corresponding to it with respective embeddings $H_v \subset H_T$. A pooled entity representation $\bar{e}_{v}  \in \mathbb{R}^d$, contextualized by sequence $T$, is computed as the mean of token embeddings $H_v$:

$$
\bar{e}_{v} = \frac{1}{|H_v|}\sum_{\bar{h}_j \in H_v} \bar{h}_j
$$

\paragraph{Subgraph Node Representations} A $d$-dimensional graph-based representation $\bar{g}_v \in \mathbb {R}^d$ for concept $v$ can be obtained by encoding local KG subgraph $\mathcal{G}_v$ with a graph encoder: $\bar{g}_v = GNN(\mathcal{G}_v)$. To obtain a KG-based vector representation $\bar{g}_v$ for $v$, we utilize a multi-layer Graph Attention Network (GAT)~\citep{DBLP:conf/iclr/VelickovicCCRLB18,DBLP:conf/iclr/Brody0Y22} that iteratively updates node representation under Message Passing framework~\citep{DBLP:conf/icml/GilmerSRVD17}:
$$
\bar{g}^{(l)}_v = \sigma \left( \sum_{(u, r, v) \,\in\,\mathcal{E}_v} \alpha^{l}_{uv} \cdot W^{l} \bar{g}^{(l-1)}_u  + W^l_o  \bar{g}^{(l-1)}_v\right)
$$
$$
\alpha^{l}_{uv} =  \frac{exp(e^{l}_{uv})}{\sum_{(w, r, v) \,\in\,\mathcal{E}_v} exp(e^{l}_{wv})}
$$
$$
e^{l}_{uv} =  a^T \cdot \sigma(W^{l} \cdot [\bar{g}^{(l-1)}_u \mathbin\Vert  \bar{g}^{(l-1)}_v]),
$$ 

where $\alpha_{uv}$ is the attention weight for an edge $(u, r, v)$, $W^l, W^l_o \in \mathbb{R}^{d \times d}$ are weight matrices, $l$ is a layer number, and $\sigma$ is a LeakyRELU activation. A graph embedding $\bar{g}^{(l)}_v$ from the last GNN encoder layer is used as a concept's graph embedding: $\bar{g}_v=\bar{g}^{(l)}_v$. Unlike prior research on KG-LM fusion~\citep{DBLP:conf/naacl/YasunagaRBLL21,DBLP:conf/nips/YasunagaBR0MLL22}, we attempt to add semantical and lexical information to graph embeddings. As an initial representation for a node $u$, a random concept name $s_u \in S_u$ is sampled and encoded with a textual encoder: $\bar{g}^{(0)}_u = LM(s_u)$. Thus, graph encoder $GNN$ is provided with additional semantics captured by the textual encoder $LM$.



\paragraph{Linearized Graph Representation} An alternative to the introduction of an additional external graph neural network is to linearize a set of graph triples into a textual graph summary encoded with an LM~\citep{DBLP:conf/aaai/LiuZ0WJD020,DBLP:conf/acl/KeJRCWSZH21,DBLP:conf/paclic/SalnikovLRNBMP23}. Since KG nodes are often attributed with textual representations (e.g., textual concept names in UMLS KG), this approach allows transferring knowledge learned from raw texts to graph representations. To obtain a linearization $L(\mathcal{G}_v)$ of graph $\mathcal{G}_v$, we linearize each edge $(u, r, v) \in \mathcal{E}_v$ as: "$L(u, r, v) = s_u\,\, r\,\,s_v\,\,[SEP]$", where $s_u \in S_u$ is a randomly sampled name of concept $u$. The resulting linearized graph obtained as the concatenation of concept name $s_v \in S_v$ and  linearized edges from  $\mathcal{E}_v$ is further encoded with a textual encoder:
$$
\bar{g}_v = LM\left( [CLS] \,\,s_v\,\,[SEP] \bigoplus_{(u, r, v) \in \mathcal{E}_v} L(u, r, v)\right),
$$
where $\bigoplus$ is a string concatenation.

\subsection{Training Objectives} \label{sec:obj}

\paragraph{Masked Language Modeling (MLM)} MLM, a widely used pretraining objective for language models, has proven effective both in the general domain~\cite{DBLP:conf/naacl/DevlinCLT19,liu2019roberta,DBLP:conf/nips/YasunagaBR0MLL22} and in the biomedical domain~\cite{DBLP:journals/health/GuTCLULNGP22,DBLP:conf/acl/YasunagaLL22,DBLP:conf/nips/YasunagaBR0MLL22}. The objective aims to make a model learn informative token representations $H_T$ by predicting masked tokens from unmasked ones using a corrupted input text as context. Specifically, given a subset of tokens $M \subset T$ masked with a masking token $[MASK]$, the model aims to restore the original tokens relying on the remaining ones as context:

$$
\mathcal{L}_{MLM} = - \sum_{t_i \in M} \log{ p(t_i | H_T)}
$$

\paragraph{Cross-Modal Alignment}

Our alignment procedure is designed to enhance a textual encoder $LM$ with domain-specific knowledge through contrastive learning using mentioned entities as anchors. Specifically, given a batch $\{(\bar{e}_i, \bar{g}_i)\}^B_{i=1} $ consisting of $B$ aligned paired text-graph representations, we introduce a  InfoNCE~\citep{Oord2018RepresentationLW} contrastive objective to pull two representations a concept $v_i$ closer in the aligned embedding space:

$$ \mathcal{L}_{align} = -\frac{1}{B}\sum^B_{i=1}\left(\log\frac{\exp(cos(\bar{e}_i,\bar{g}_i)/\tau)}{\sum^B_{j=1}\exp(cos(\bar{e}_j,\bar{g}_j)/\tau)}\right), $$


where $B$ is a batch size, and $\tau > 0$ is a temperature parameter, and $cos(\bar{e}_i,\bar{g}_i)$ is a cosine similarity between $\bar{e}_i$ and $\bar{g}_i$. Since the entity representation $\bar{e}_i$ is pooled over a textual sequence masked for MLM objective, alignment loss further enforces LM to infer relevant information from the whole sequence $T$.

The resulting loss is a sum of MLM and alignment objective: 
$$\mathcal{L} = \mathcal{L}_{MLM} + \mathcal{L}_{align}$$ 
Intuitively, the training objective is designed to encourage an LM enrich entity representation with external knowledge from a KG while retaining its language understanding through continuous MLM pre-training.

After the BALI pretraining, the GNN module gets discarded and only the resulting language encoder $LM$ is used for downstream tasks. Here, our assumption is that $LM$ obtains additional factual domain-specific knowledge from a reference KG.

\begin{table}[ht!]
\setlength{\tabcolsep}{2pt}
\centering
\caption{Hyperparameter values used for BALI pretraining. LR stands for learning rate.}\label{tab:hyperparams}
\begin{tabular}{l|cc}
\textbf{Hyperparameter} & \textbf{Base models}  & \textbf{Large models}\\  \midrule
GNN hidden size & 768 & 768 \\
Max \# of node neighbors  & 3 & 3  \\
\# of graph encoder layers & 5 & 5  \\
GAT's \# of attention heads & 2 & 2 \\

LM parameters LR & $2 \cdot 10^{-5}$ & $1 \cdot 10^{-5}$ \\
Non-LM parameters LR & $1 \cdot 10^{-4}$ & $1 \cdot 10^{-4}$ \\
Batch size & 256 & 256 \\ 
\# of epochs & 10 & 10 \\
\# of steps & 65,000 & 65,000 \\

\bottomrule
\end{tabular}
\end{table}

\section{Experiments}


To assess the effectiveness of our proposed methodology, we first pre-train existing biomedical LMs with the BALI method and then assess the performance of the resulting models in various biomedical NLP tasks.






\begin{table}[t]
\setlength{\tabcolsep}{2.pt}
\caption{Mean evaluation accuracy and standard deviation across 10 evaluation runs for proposed BALI alignment procedure on biomedical question answering datasets. \emph{GNN} stands for BALI with external GAT graph encoder while \emph{Linear graph} stands for single-encoder implementation with KG subgraphs encoded with an LM. }
\label{tab:res_qa}
\centering
\begin{tabular}{l|ccc}
\textbf{Model} & \textbf{PubMedQA} & \textbf{MedQA} & \textbf{BioASQ 2023} \\
\midrule
PubMedBERT$_{base}$ & 63.1 $\pm$ 2.9  & 38.1  & 67.8 $\pm$ 4.1   \\
\,\,\,+ BALI (GNN) & \underline{65.2} $\pm$ 1.2  & \underline{39.8}  & \underline{74} $\pm$ 3.4  \\
\,\,\,+ BALI (Linear graph) & 65.0 $\pm$ 1.6 & 38.8 & 72.2 $\pm$ 4.4 \\
\midrule
BioLinkBERT$_{base}$ & 63.3 $\pm$ 3.6  & 40.0  & 65.9 $\pm$ 2.7 \\
\,\,\,+ BALI (GNN) & \underline{64.4} $\pm$ 2.1  & \underline{43.1}  & \underline{73.6} $\pm$ 3.6  \\
\,\,\,+ BALI (Linear graph) & 63.9 $\pm$ 4.4  & 40.5 & 65.7 $\pm$ 3.6   \\


\midrule
BioLinkBERT$_{large}$ & 69.5 $\pm$ 2.4 & 44.6 & 67.7 $\pm$ 3.7 \\

\,\,\,+ BALI (GNN)  & 68.7 $\pm$ 5.2 & 45.0 & \underline{67.9} $\pm$ 4.5 \\
\,\,\,+ BALI (Linear graph)  & \underline{70.9} $\pm$ 1.7 & \underline{45.5} & 66.0 $\pm$ 6.1 \\\midrule \multicolumn{4}{c}{\textbf{Task-specific joint LM-KG reasoning QA methods}} \\\midrule
QA-GNN~\citep{DBLP:conf/naacl/YasunagaRBLL21} & 72.1 & 45.0 & --- \\
GreaseLM~\citep{DBLP:conf/iclr/0001BYRLML22} & 72.4 & 45.1 & --- \\
DRAGON~\citep{DBLP:conf/nips/YasunagaBR0MLL22} & 73.4 & 47.5 & --- \\

\midrule \multicolumn{4}{c}{\textbf{Large Language Models}} \\\midrule
GPT-4~\citep{DBLP:journals/corr/abs-2303-13375} & 80.4 & 86.1 & 95.4 \\
    
\bottomrule
\end{tabular}





\vspace{-0.2cm}
\end{table}






\paragraph{Pretraining Data} 
As pretraining data, we adopt the PubMed abstracts\footnote{\url{pubmed.ncbi.nlm.nih.gov/}} with biomedical entities recognized and normalized to the UMLS KG (version 2020AB) with the BERN2 tool~\citep{10.1093/bioinformatics/btac598}.  Given the substantial entity distribution imbalance in scientific abstracts, with entities like human, mice, and cancer being the most common ones, we address this issue as follows. To ensure a more balanced dataset with diverse concepts, we sample only up to 10 sentences from PubMed abstracts iteratively for each concept present in the UMLS. The resulting dataset has 1.67M sentences with mentioned entities covering about 600K unique UMLS concepts.

\subsection{Evaluation Tasks} We evaluate the effectiveness of our proposed  alignment method on the following knowledge-demanding tasks in biomedical domain:

\paragraph{Question Answering (QA)}
To assess how the entity-level representation alignment influences an LM's overall factual awareness on biomedical domain-specific knowledge, we adopt three  QA datasets: (i) PubMedQA~\citep{DBLP:conf/emnlp/JinDLCL19}; (ii) MedQA-USMLE~\citep{app11146421}; (iii) BioASQ 2023~\citep{DBLP:conf/clef/NentidisKKP23}. We employ accuracy as the evaluation metric.
    
\paragraph{Entity Linking (EL)} For biomedical entity linking, we adopt 5 corpora: (i) NCBI~\cite{NcbiCorpus}, (ii) BC5CDR-D~\cite{BC5CDRcorpus}, (iii) BC5CDR-D~\cite{BC5CDRcorpus}, (iv) BC2GN~\cite{morgan2008overview}, (v) SMM4H~\cite{SMM4HCorpus2017}. We consider two scenarios: (i) zero-shot similarity-based retrieval approach over pooled mention and concept name representations~\citep{FairEvaluationInConceptNormalization}; (ii) supervised approach based on BioSyn~\citep{sung-etal-2020-biomedical-biosyn}, a model that
iteratively updates candidates list using synonym marginalization. Following prior EL research~\citep{DBLP:conf/acl/PhanST19,sung-etal-2020-biomedical-biosyn,FairEvaluationInConceptNormalization,sakhovskiy-etal-2024-biomedical}, we employ the top-$k$ accuracy as the evaluation metric: $\mathrm{Acc@k}=1$ if the correct concept is retrieved at the rank 
$\le k$, otherwise $\mathrm{Acc@k}=0$.

\paragraph{Relation Extraction} Additionally, we perform evaluation on three biomedical relation extraction datasets: (i) Chemical Protein Interaction corpus (ChemProt)~\citep{ChemProtDataset2017}, (ii) Drug-Drug Interaction corpus (DDI)~\citep{DDICorpus}, and (iii)  Genetic Association Database (GAD)~\citep{GADCorpus}. For evaluation, we adopt the micro-averaged F1-score following prior research~\citep{DBLP:journals/health/GuTCLULNGP22,ChatGPTOnBLURB}. We compare against domain-specific language encoders and  ChatGPT evaluation results from~\citet{ChatGPTOnBLURB}.

\subsection{Evaluation Datasets}\label{appx:datasets}

\emph{PubMedQA}~\citep{DBLP:conf/emnlp/JinDLCL19} is a collection of 1,000 3-way question answering dataset with questions derived from PubMed abstracts. Each question has a single correct answer from yes/no/maybe options.

\emph{MedQA}~\citep{app11146421} provides multiple choice questions derived from the US National Medical Board Examination, each offering 4 answer choices. We adopt the English part of the corpus which offers 12,723 unique questions.

\emph{BioASQ 2023}~\citep{DBLP:conf/clef/NentidisKKP23} is a part of BioASQ shared tasks series on biomedical natural language processing. BioASQ has 1,357 binary yes/no questions manually curated by experts in the biomedical domain.

\emph{The NCBI Disease Corpus}~\cite{NcbiCorpus} contains 793 PubMed abstracts with disease mentions and their concepts corresponding to the MEDIC dictionary~\citep{davis2012medic}. It has 5134, 787, and 204 entities in train, dev, and test set after filtration of simple cases such as train-test and dictionary-test set intersection, respectively.

\emph{BC5CDR}~\citep{BC5CDRcorpus}  provides a task for the extraction of chemical-disease relations (CDR) from 1500 PubMed abstracts that contains annotations of both chemical/diseases. The disease part has 4182, 4244, and 657 entities in train, dev, and test set after filtration, respectively. The chemical part contains 5203, 5347, and 425 entities, respectively.

 \emph{BioCreative II GN}~\citep{morgan2008overview}  contains PubMed abstracts with human gene and gene product mentions for gene normalization (GN) to Entrez Gene identifiers~\citep{maglott2007entrez}. There are 2,725/985 train/test entities.

The Social Media Mining for Health (\emph{SMM4H}) challenge~\citep{SMM4HCorpus2017} released a dataset with annotated ADR mentions linked to MedDRA. Tweets were collected using 250 generic and trade names for therapeutic drugs. Manually extracted ADR expressions were mapped to Preferred Terms (PTs) of the MedDRA dictionary. The dataset provides 6650/831 train/test entities.

The Chemical Protein Interaction corpus (\emph{ChemProt})~\citep{ChemProtDataset2017} covers chemical-protein interactions between chemical and protein entities extracted from PubMed abstracts. In total, there are 23 interaction types. The dataset includes 18035/11268/15745 samples in train/dev/test sets.

\emph{DDI}~\citep{DDICorpus} is a Drug-Drug Interaction corpus designed for research on pharmaceutical information
extraction. It consists sentence-level annotations for drug-drug interactions from PubMed abstracts. The corpus has 25296/2496/5716 train/dev/test samples.

\emph{GAD} is the semi-automatically collected Genetic Association Database corpus of gene-disease interactions from PubMed abstracts. It has 4261/535/534 samples in train/dev/test.

\begin{table*}[t]
\caption{Evaluation results on biomedical entity linking in zero-shot and supervised set-ups. @1 and @5 stand for Accuracy@1 and Accuracy@5, respectively. For each model, underline highlights the best of two scores: (i) retrieval accuracy of the original biomedical LM and (ii) the score for model pre-trained with BALI method. }
\label{tab:res_linking}
\centering
\begin{tabular}{l|cccccccccc}

\multirow{2}{*}{\textbf{Model}} & \multicolumn{2}{c}{\textbf{NCBI}}  & \multicolumn{2}{c}{\textbf{BC5CDR-D}} & \multicolumn{2}{c}{\textbf{BC5CDR-C}} & \multicolumn{2}{c}{\textbf{BC2GM}} & \multicolumn{2}{c}{\textbf{SMM4H}}  \\

\cmidrule(lr){2-3} 
\cmidrule(lr){4-5} \cmidrule(lr){6-7}
\cmidrule(lr){8-9}
\cmidrule(lr){10-11}

 & @1 & @5 & @1 & @5 & @1 & @5 & @1 & @5 & @1 & @5 \\ \hline 

\multicolumn{11}{c}{\textbf{Zero-shot evaluation, general-purpose biomedical LMs}} \\
\hline 

PubMedBERT & 49.51 & 65.69 & 58.75 & 75.04 & 76.24 & 80.24 & 68.12 & 74.11 & 16.13 & 25.27 \\
\,\,\,+ BALI & \underline{68.14} & \underline{79.90} & \underline{72.30} & \underline{81.28} & \underline{85.65} & \underline{89.65} & \underline{83.25} & \underline{89.44} & \underline{24.91} & \underline{36.82} \\\midrule

BioLinkBERT$_{base}$ & 35.78 & 44.12 & 45.81 & 54.64 & 70.59 & 73.41 & 58.17 & 61.52 & 8.30 & 10.83  \\

\,\,\,+ BALI & \underline{68.63} & \underline{78.92} & \underline{73.82} & \underline{82.65} & \underline{86.59} & \underline{90.82} & \underline{82.64} & \underline{89.24} & \underline{27.92} & \underline{43.08}   \\\midrule

BioLinkBERT$_{large}$ & 32.35 & 42.65 & 44.29 & 50.99 & 70.12 & 73.18  & 57.66 & 62.13 & 8.54 & 12.27 \\
\,\,\,+ BALI & \underline{70.1} & \underline{78.92} & \underline{73.21} & \underline{80.67} & \underline{85.65} & \underline{90.12} & \underline{82.44} & \underline{89.04} & \underline{22.98} & \underline{34.78}  \\\midrule

\multicolumn{11}{c}{\textbf{Zero-shot evaluation, task-specific EL Models}} \\\hline 

SapBERT & \underline{71.57} & \underline{84.31} & 73.67 & \underline{84.32} & 85.88 & \underline{91.29} & \underline{87.61} & \underline{92.18} & \underline{39.59} & \underline{58.84}  \\
\,\,\,+ BALI & \underline{71.57} & 81.86 & \underline{74.28} & 82.50 & \underline{86.35} & 90.35 & 85.89 & 91.37 & 28.04 & 42.00  \\\midrule

GEBERT & 70.59 & \underline{83.33} & \underline{74.58} & \underline{85.39} & 85.41 & 91.76 & \underline{87.21} & \underline{92.79} & \underline{38.27} & \underline{62.33} \\
\,\,\,+ BALI & \underline{73.04} & 81.86 & 73.52 & 82.50 & \underline{86.59} & \underline{92.0} & 85.48 & 91.57 & 28.04 & 46.21 \\


\hline
\multicolumn{11}{c}{\textbf{Supervised evaluation, general-purpose biomedical LMs}} \\
\hline

PubMedBERT & 72.06 & 84.31 & \underline{74.73} & \underline{83.71} & 86.12 & 92.00 & 87.92 & \underline{92.39} & 66.19 & \underline{79.90}   \\
\,\,\,+ BALI & \underline{74.02} & \underline{82.35} & \underline{74.73} & 81.74 & \underline{87.76} & \underline{92.94} & \underline{88.32} & 91.88 & \underline{68.71} & 79.66  \\\midrule
BioLinkBERT$_{base}$ & 56.86 & 70.59 & 74.58 & \underline{85.39} & 87.29 & \underline{92.94} & \underline{88.32} & 92.39 & 65.94 & 77.74  \\
\,\,\,+ BALI & \underline{75.00} & \underline{84.31} & \underline{75.49} & 83.26 & \underline{88.94} & 92.71 & \underline{88.32} & \underline{92.89} & \underline{67.27} & \underline{78.34}  \\\midrule

BioLinkBERT$_{large}$ & 57.35 & 70.10 & 65.91 & 77.32  & 79.53 & 86.82 & 76.95 & 88.43 & 38.99  & 64.02  \\
\,\,\,+ BALI & \underline{74.51} & \underline{82.35} & \underline{74.28} & \underline{82.34} & \underline{87.53} & \underline{93.88} & \underline{88.73} & \underline{92.79} & \underline{66.06} & \underline{79.06} \\\midrule

\multicolumn{11}{c}{\textbf{Supervised evaluation, task-specific biomedical EL models}} \\\hline 

SapBERT & \underline{75.00} & \underline{85.78} & 74.58 & \underline{84.47} & 86.59 & \underline{93.18} & \underline{89.24} & \underline{93.71} & 66.79 & \underline{80.51} \\
\,\,\,+ BALI & 74.51 & 83.82 & \underline{74.73} & 82.80 & \underline{88.24} & \underline{93.18} & 88.12 & 92.79 & \underline{69.19} & 78.94 \\\midrule

GEBERT & 73.04 & \underline{84.80} & \underline{75.80} & \underline{85.39} & 87.06 & 92.71 & \underline{88.83} & \underline{93.71} & 65.70 & 80.63  \\
\,\,\,+ BALI & \underline{74.02} & 83.33 & 75.49 & 83.87 & \underline{89.41} & \underline{93.65} & 88.22 & 93.50 & \underline{67.51} & \underline{80.75} \\


\bottomrule

\end{tabular}
\end{table*}

    


\paragraph{Pre-training set-up \& Implementation Details.} We trained our models for 65k steps (10 epochs) with a batch size of 256 using AdamW~\citep{DBLP:conf/iclr/LoshchilovH19} optimizer with a peak learning rate of $2\cdot 10^{-5}$ for LM parameters and $1 \cdot 10^{-4}$ for other parameters and cosine learning rate decay to zero. Following~\cite{DBLP:conf/nips/HamiltonYL17}, we sample a subset of up to 3 neighboring nodes to reduce computational cost of our model. Higher neighbors count as well as the extension to 2- and 3-hop neighbors would result in a significant memory footprint growth as well as excessive training time. For MLM objective, we follow the original set-up proposed in BERT~\citep{DBLP:conf/naacl/DevlinCLT19} by selecting $15\%$ of input tokens.  Each selected token is either replaced with a special $[MASK]$ token, left unchanged, or replaced by a randomly selected vocabulary token with probabilities of $0.8$, $0.1$, and $0.1$, respectively. As base models, we adopt PubMedBERT\footnote{huggingface:microsoft/BiomedNLP-BiomedBERT-base-uncased-abstract-fulltext}~\citep{DBLP:journals/health/GuTCLULNGP22} and BioLinkBERT\footnote{huggingface:michiyasunaga/BioLinkBERT-base}\footnote{huggingface:michiyasunaga/BioLinkBERT-large}~\citep{DBLP:conf/acl/YasunagaLL22}, state-of-the-art biomedical LMs that are pre-trained on scientific articles from PubMed. In our experiments, we pre-train each \emph{base}- and \emph{large}-sized BALI model for 65K with a batch size of 256. Table~\ref{tab:hyperparams} lists hyperparameter values used for pre-training of BALI models.

\paragraph{Hardware \& Software set-up} All models in our experiments were trained and evaluated using  the version 1.11.0 of PyTorch~\cite{NEURIPS2019_9015} with CUDA 11.3~\cite{nickolls2008scalable} support. GAT~\citep{DBLP:conf/iclr/Brody0Y22} and GraphSAGE~\citep{DBLP:conf/nips/HamiltonYL17} graph neural networks were adopted from the PyTorch Geometric~\cite{Fey/Lenssen/2019} library (version 2.0.4). The pretraining of each base-sized BALI model took approximately 9 hours on 4 NVIDIA V100 GPUs and 8 CPU cores. The pretraining of large-sized BALI models took 10 hours on 8 NVIDIA V100 GPUs and 16 CPU cores. For both base and large models we adopt ZeRO~\citep{ZeROOptimization2022} stage 2 from Deepspeed~\cite{DBLP:conf/kdd/RasleyRRH20}. All evaluation experiments were conducted on  a machine with single NVIDIA V100 GPU.



\paragraph{Evaluation set-up} To explore the effectiveness of BALI, we compare each pre-trained alignment model against its base model with the original weights. Notably, PubMedBERT and BioLinkBERT models are also trained on scientific texts from PubMed database and only differ in pre-training objective. Additionally, we compare against strong task-specific baselines as well as general-purpose Large Language Models (LLMs). Due to small dataset sizes and fine-tuning instability, we average performance across 10 runs on PubMedQA and BioASQ corpora.

\paragraph{QA baselines} For QA, we employ task-specific QA-GNN~\citep{DBLP:conf/naacl/YasunagaRBLL21} and GreaseLM~\citep{DBLP:conf/iclr/0001BYRLML22} models that enhance backbone BioLinkBERT$_{large}$ with relevant UMLS KG subgraph as well as reasoning module available during inference time. Unlike BALI, these methods implement the inference-time LM-KG interaction. DRAGON~\citep{DBLP:conf/nips/YasunagaBR0MLL22} extends QA-GNN and GreaseLM by introducing bidirectional LM-KG interaction for both knowledge-enhanced pre-training and fine-tuning. For PubMedQA and MedQA, we compare with 5-shot GPT-4 prompting~\citep{DBLP:journals/corr/abs-2303-13375}. For BioASQ, we adopt the zero-shot prompting approach~\cite{DBLP:conf/clef/HsuehZLHMT23} as it has proved an effective approach in the official BioASQ 2023 Shared Task results overview~\citep{BioASQ2023SharedTask}.

\paragraph{EL baselines} For entity linking, we adopt SapBERT\footnote{huggingface:cambridgeltl/SapBERT-from-PubMedBERT-fulltext}~\citep{SapBERTMonoLingual2022} and GEBERT\footnote{huggingface:andorei/gebert\_eng\_gat/}~\citep{DBLP:conf/clef/SakhovskiySKT23} which are a PubMedBERT additionally pre-trained for synonymous concept name clusterization objective on all concepts available in the UMLS KG. GEBERT additionally performs concept clusterization in node representation space followed by representation alignment between textual and graph encoders. 


\begin{table}[t]
\caption{Evaluation results on biomedical relation extraction in terms of Micro F1. For each model, \underline{underline} highlights the best quality among the original biomedical LM and model pre-trained with BALI method. The best results for each dataset are highlighted in \textbf{bold}. }
\label{tab:res_rel_extraction}
\centering
\setlength{\tabcolsep}{3.5pt}
\begin{tabular}{l|ccc}

\textbf{Model} & \textbf{ChemProt} & \textbf{DDI} & \textbf{GAD}  \\
\midrule

PubMedBERT & 76.57 & 79.02 & 83.36  \\
\,\,\, + BALI & \underline{76.91} & \textbf{\underline{81.17}} & \textbf{\underline{83.68}} \\\midrule
BioLinkBERT$_{base}$ & 76.97 & \underline{79.79} & 81.97  \\
\,\,\, + BALI & \textbf{\underline{77.52}} & 79.69 & \underline{82.71} \\\midrule
SapBERT & \underline{76.74} & 80.09 & 80.94  \\
\,\,\, + BALI & 76.47 & \underline{80.56} & \underline{82.12} \\\midrule
GEBERT & 75.31 & 80.01 & 80.73  \\
\,\,\, + BALI & \underline{77.32} & \underline{80.95} & \underline{83.33} \\\midrule
ChatGPT~\citep{ChatGPTOnBLURB} & 34.16 & 51.62 & 52.43  \\
\bottomrule

\end{tabular}
\end{table}

\paragraph{Evaluation Implementation} For fine-tuning on QA and relation extraction experiments, we adopt the hyperparameters from BioLinkBERT~\citep{DBLP:conf/acl/YasunagaLL22} for better comparability of the experimental results. The main difference is that we load the model weights from the best epoch in terms of dev set quality metric. For zero-shot linking, we adopt the retrieval code from~\cite{tutubalina2020fair}. For BioSyn~\citep{sung-etal-2020-biomedical-biosyn}, we adopt the default hyperparameters and train each model for 20 epochs with a learning rate of $1 \cdot 10^{-5}$ and a batch size of 16.


\subsection{Results}

To answer the \textbf{RQ 1}, we assess our methodology on biomedical QA and entity linking.

\subsubsection{Results: Question Answering}

The evaluation results for pre-trained BALI models on biomedical QA datasets are presented in Table~\ref{tab:res_qa}. Across all datasets, BALI consistently boosts baseline models, for instance, PubMedBERT aligned through an external GAT encoder demonstrates 2.1\%, 1.7\%, and 6.2\% mean accuracy gain on PubMedQA, MedQA, BioASQ, respectively. Interestingly, the adoption of an external graph encoder (GNN) gives more improvement than a single-encoder set-up with linearized graph for \emph{base}-sized models while BioLinkBERT$_{large}$ achieves higher accuracy in linearized setting rather than GNN-based setting. Apparently, small language encoders do not have enough capacity to simultaneously encode two modalities while larger model can capture more domain-specific knowledge through the BALI representation alignment procedures. 

Despite BioLinkBERT$_{large}$ has no access to a retrieved KG subgraph for inference-time reasoning, after BALI pretraining it performs on par or better than the task-specific QA-GNN and GreaseLM methods that reason over retrieved KG subgraphs.
In contrast to state-of-the-art retrieval-augmented encoder-based methods for KGQA, the effective integration of an auxiliary Knowledge Graph (KG) modality during pre-training enables encoders pre-trained with BALI to effectively handle knowledge-demanding questions. This capability persists even in the absence or failure of a KG retrieval system at runtime.
Notably, both QA-GNN and GreaseLM have BioLinkBERT$_{large}$ as a backbone LM. However, there is a huge performance gap between small encoders and GPT-4.

\subsubsection{Results: Entity Linking}

Table~\ref{tab:res_linking} presents the evaluation results for models aligned using an extrenal GNN on entity linking in both zero-shot and supervised settings. As seen from the results, BALI increases entity linking capabilities of general-purpose biomedical LMs in both set-ups, especially in zero-shot settings. For instance, PubMedBERT and BioLinkBERT$_{base}$ show huge average Accuracy@1 gains of 13.1\% and 24.2\% across all datasets in zero-shot evaluation, respectively. As specialized SapBERT and GEBERT models were trained on synonymous concept names from full UMLS knowledge base, they exhibit strong zero-shot robustness to lexical variability of SMM4H's social media domain. BALI pretraining reduces the linking quality by about 10\% Accuracy@1 for both models. Cross-modal alignment pretraining seems to affect the clustered structure of these models' representation space negatively. However, in the academic domain of the four remaining corpora, BALI pretraining does not reduce the zero-shot capabilities of specialized EL models significantly, and it even results in a slight Accuracy@1 increase on the NCBI and BC5CDR-C corpora.

\begin{table}[t]
\caption{Ablation analysis for BALI model with PubMedBERT base model. For each ablation-set-up and dataset, mean accuracy across 10 runs with different random states is reported.}
\label{tab:res_graph_encoder_choice}
\centering
\begin{tabular}{l|cc}
\setlength{\tabcolsep}{2pt}
\textbf{Node representation} & \textbf{PubMedQA} &  \textbf{BioASQ} \\
\midrule
\multicolumn{3}{c}{\textbf{Double encoder set-ups}} \\\hline 

GNN (GAT) & 65.2  & \textbf{74} \\
GNN (GraphSAGE) & 58.8   & 70 \\\hline
\multicolumn{3}{c}{\textbf{Single LM encoder set-ups}} \\\hline 
LM + Linear graph & 65.0  & $72.2$ \\
LM + DistMult & \textbf{65.44}  & 69.77 \\
LM + TransE & 64.22  & 70.47 \\
Textual & 58.86  & 71.40  \\
\bottomrule
\end{tabular}
\end{table}



\begin{table}[t]

\caption{Ablation analysis for BALI model with PubMedBERT base model and GAT graph encoder. For each ablation-set-up and dataset, mean accuracy across 10 runs with different random states is reported. }
\label{tab:res_ablation}
\centering
\begin{tabular}{l|cc}
\setlength{\tabcolsep}{1.75pt}
\textbf{Model} & \textbf{PubMedQA} &  \textbf{BioASQ 2023} \\
\midrule
PubMedBERT & 63.1   & 67.8   \\
\,\,+ BALI & \textbf{65.2}   & \textbf{74} \\\midrule
\multicolumn{3}{c}{\textbf{Training objective}} \\\midrule
$-\mathcal{L}_{MLM}$ & 60.54  & 64.53 \\
$-\mathcal{L}_{align}$ & 63.78  & 70.58 \\
$-\mathcal{L}_{align}, +\mathcal{L}_{ms-align}$ & 62.60  & 72.44 \\
$-\mathcal{L}_{align}$, + classification & 63.52 & 72.56 \\
\midrule
\multicolumn{3}{c}{\textbf{Token-entity aggregation}} \\\midrule
Weighted & 63.20 & 
 70.58 \\
GAT & 64.50  & 70.58 \\ 
Transformer layer & 63.06  & 71.40  \\\midrule
\multicolumn{3}{c}{\textbf{\# Graph encoder layers}} \\\midrule
$L=3$ & 64.72  & 71.51 \\
$L=7$ & 62.62  & 69.07 \\
\bottomrule
\end{tabular}
\end{table}

After supervision, general-purpose biomedical LMs pre-trained with BALI achieve retrieval accuracy comparable to state-of-the-art biomedical EL methods. For instance, BioLinkBERT$_{base}$ with BALI pretraining performs on par or slightly better than the task-specific SapBERT model that is pretrained on all synonyms available in UMLS on 2 of 5 corpora (namely, BC5CDR-Disease and BC5CDR-Chem). The influence of BALI on academic datasets is similar to zero-shot set-up: alignment does not give a consistent improvement over baseline model. Interestingly, SapBERT and GEBERT obtain an Accuracy@1 improvement of 2.4\% and 1.8\%, respectively, on SMM4H while showing a huge performance drop in zero-shot evaluation. Seemingly, BALI enriches embeddings with additional information distilled from KG modality and produces distinguishable and informative biomedical concept representations.
\subsubsection{Results: Relation Extraction}

The evaluation results for relation extraction using GNN-based alignment are presented in Table~\ref{tab:res_rel_extraction}. Due to computational instability, we do not report the evaluation results for BioLinkBERT$_{large}$. As seen from the results, BALI representation alignment improves relation extraction on average with only exceptions being SapBERT (-0.3\% Micro F1) on ChemProt and BioLinkBERT on DDI (-0.1\% Micro F1). The best results for all three datasets are achieved after BALI pretraining with 0.6\%, 1.1\%, and 0.3\% Micro F1 increase over the best non-aligned model on ChemProt, DDI, and GAD corpora. All the baselines and BALI models outperform LLM-based approach by a large margin highlighting the importance of specialized models for domain-specific information retrieval applications.



\subsection{Analysis: Node Representation}

To answer the \textbf{RQ 2} and \textbf{RQ 3}, we implement BALI with different graph representation methods. Under the GNN approach, we pre-train and evaluate BALI implementation with GraphSAGE~\citep{DBLP:conf/nips/HamiltonYL17} instead of GAT which adopts mean-pooling instead of attention aggregation across neighboring nodes.

\paragraph{Translation-based Node Representations} In a series of graph representation methods~\citep{DBLP:journals/corr/YangYHGD14a,DBLP:conf/nips/BordesUGWY13,DBLP:conf/icml/TrouillonWRGB16,DBLP:conf/iclr/SunDNT19}, a relation triplet (graph edge) $(u, r, v) \in \mathcal{E}$ is modeled as a relation-based translation of the head node $v$ with a relational transformation $f_r$: $u \approx f_r(v)$. Similarly to linearized graph set-up, a single LM is used to encode both head and tail nodes of each edge. The only parameters introduced for translation-based node representations are contained within relation-specific translations. In our work, we adopt DistMult~\cite{DBLP:journals/corr/YangYHGD14a} and TransE~\cite{DBLP:conf/nips/BordesUGWY13} to represent in-context entity representation $\bar{e}_v$ as a transformation of concept name embedding $\bar{g}_u = f_r(LM(s_u))$ for $s_u \in S_u$. 


\paragraph{Textual Node Representations} To assess the necessity of a graph encoder for capturing additional information not accessible to the language encoder LM, we perform experiments using node embeddings that rely exclusively on textual concept names.  In particular, we compute a node embedding $\bar{g}_u$ by mean pooling the textual output of a randomly chosen concept name $s_u \in S_u: \bar{g}_u = LM(s_u)$.

\paragraph{Analysis: Node Representation Choice} Experiments with different node representation types are summarized in Table~\ref{tab:res_graph_encoder_choice}. Based on the results, we can make the following observations. First, simpler mean-pooling local subgraph aggregation under the GNN-based approach leads to a significant performance drop of 6.4\% and 4\% on PubMedQA and BioASQ which highlights the importance of learning relative node importance scores: not all nodes are equally useful. Despite its simplicity, translation-based DistMult and TransE models show high performance in our alignment procedure in combination with LM. Similarly, a linearized graph encoded with LM seems to be the closest to BALI implementation with a GAT encoder. Thus, we conclude that LMs can serve as an effective graph representation method for text-attributed graph for LM-KG alignment. Finally, textual node representations with no KG subgraph provided have shown poor performance indicating the performance of additional local graph context for text-graph alignment.

\subsection{Ablation Study}

To justify modeling choices made in BALI model, we perform an extensive analysis in three directions: (i) Training loss choice, (ii) Token-entity aggregation, (iii) graph encoder size.  For each ablation, we pre-train a separate BALI model with PubMed initialization and summarize evaluation results across $10$ runs with different random states on PubMedQA and BioASQ. The results are summarized in Table~\ref{tab:res_ablation}.

Although the BALI implementation with InfoNCE contrastive loss has shown promising results, we conducted additional experiment with Multi-Similarity contrastive loss (MS-loss)~\citep{wang2019multi} as an alternative to $\mathcal{L}_{align}$. MS-loss was shown to be effective for learning UMLS-based concept representations in prior research~\citep{SapBERTMonoLingual2022,SapBERTMultilingual,DBLP:journals/jbi/YuanZSLWY22}  including cross-modal joint textual and graph concept representations~\citep{DBLP:conf/clef/SakhovskiySKT23,sakhovskiy-etal-2024-biomedical}. The implementation of cross-modal MS loss alignment loss $\mathcal{L}_{ms-align}$ is formalized as follows:  


\begin{equation}\label{eqn:msloss}
\begin{split}
\mathcal{L}_{ms-align} = \frac{1}{|B|} \sum^{|B|}_{v=1} 
\left( 
    \frac{1}{\alpha} \log\Bigl(1 + \sum_{n \in \mathcal{N}_v} e^{\alpha(S_{vn} - \epsilon)} \Bigr) 
    + \right. \\
    + \left. 
    \frac{1}{\beta} \log \Bigl( 1 + \sum_{p \in \mathcal{P}_v} e^{-\beta(S_{vp} - \epsilon)} \Bigr) 
\right),
\end{split}
\end{equation}

where $\alpha, \beta$, and $\epsilon$ are hyper-parameters. $\mathcal{P}_v$ and $\mathcal{N}_v$ are the sets of positive and negative samples for a chosen anchor entity $v$. $S_{vn}$ and $S_{vp}$ are the cosine similarities of anchor entity embedding $\bar{e}_v$ to negative graph embedding $\bar{g}_n$ and positive graph $\bar{g}_p$, respectively. We adopt the hyper-parameter values from prior biomedical concept representation research~\citep{SapBERTMonoLingual2022,DBLP:conf/clef/SakhovskiySKT23}.

\subsubsection{Ablation: training objectives} During training objective ablation study, we remove each of two losses, namely $\mathcal{L}_{MLM}$ and $\mathcal{L}_{align}$. Additionally, we attempt to replace contrastive $\mathcal{L}_{align}$ with a binary cross-entropy loss and MS-loss based contrastive alignment loss $\mathcal{L}_{ms-align}$. Specifically, given a concatenation of an entity representation $\bar{e}_i$ and a node representation $\bar{g}_j$, a fully connected 1-layer classification network is trained to determine (label $1$) whether the two embeddings represent the same concept or not (label $0$). Intuitively, this indirect classification-based representation alignment still aims to inform LM and KG encoder that two embeddings complementary describe a single biomedical concept. However, it lacks explicit information exchange between cross-modal representations pairs.

As seen from the results in Table~\ref{tab:res_ablation},  a removal of each of two losses drops the QA quality indicating that performing token-level LM-KG alignment without language modeling objectives leads to degraded language understanding. However, optimization of MLM objective alone is less effective than BALI's joint multi-task objective. Although classification-based alignment causes 0.4\% and 5.8\% classification accuracy growth on PubMedQA and BioASQ corpora compared to backbone PubMedBERT model, it falls short of contrastive BALI pretraining by 1.7\% and 1.4\%, respectively. The MS-loss BALI implementation loses 2.6\% and 1.6\% accuracy compared to the InfoNCE implementation on PubMedQA and BioASQ, respectively. Probably, this observation can be explained by the higher complexity of MS-loss in terms of and a thorough hyper-parameter search could mitigate the performance loss.

\subsubsection{Ablation: entity aggregation}  As token-entity aggregation method we experiment with following set-ups: (i) \emph{weighted} aggregation which attention weights to sum token embeddings of the last LM layer with no additional transformations; (ii) \emph{GAT} aggregation adopts single GAT layer as described in Section~\ref{sec:repres}; (iii) Transformer layer over tokens to correspond to the same entity only. All 3 token-entity aggregation alternatives perform worse than baseline BALI with mean pooling aggregation. Weighted and Transformer ablations show even lower accuracy than $-\mathcal{L}_{align}$ ablation on the PubMedQA  corpus which is equivalent to BERT pretraining without any representation alignment. Thus, entities' token representations benefit the most when treated equally.

\subsubsection{Ablation: GNN size} To explore graph encoder's capacity on graph representations quality and representations alignment, we experiment with smaller (3 layers) and larger (5 layers) GAT encoder for KG modality. Pretraining with smaller GNN encoder still maintains improvement over both the original PubMedBERT checkpoint and $-\mathcal{L}_{align}$ ablation.

\section{Discussion}

As shown in our experiments, a relatively cheap pretraining with the proposed BALI method is able to slightly improve a language encoder's effectiveness in the specialized biomedical domain. In particular, it can slightly improve the QA performance with a significant contextualized entity embedding quality improvement indicated with linking accuracy growth. These findings let us answer positively to \textbf{RQ1}, i.e., our cross-modal can be beneficial for downstream biomedical NLP tasks. In our experiments, GNN-based concept node representations tend to be more effective for cross-modal representation alignment of smaller models while linearization-based approaches can be more effective for larger encoders due to their higher parametric capacity (\textbf{RQ2}). However, a simple GNN, such as GraphSAGE, can drop the representation quality substantially compared to a stronger  GAT encoder with attention-based graph context aggregation (\textbf{RQ3}).

\section{Conclusion}

We propose BALI, a novel self-supervised pretraining method for Knowledge Graph (KG) and Language Model (LM) alignment. Experimental results indicate that the alignment of biomedical LMs enhances their performance on both question answering and entity linking tasks in the biomedical domain after a short pre-training on 1.7M sentences only. Comparison of various graph representation methods has revealed the effectiveness of both LM-based approaches with linearized graphs as well as sparse graph neural networks for capturing vital KG context absent in raw texts.
For future work, we aim to expand and apply our pre-training method to general domains and other LM architectures, such as decoder-only and encoder-decoder models.

\section*{Acknowledgements}
The work was supported by ISP RAS Research Center for Trusted Artificial Intelligence.

\bibliographystyle{ACM-Reference-Format}
\bibliography{sample-base}

\appendix

\end{document}